\documentclass[runningheads]{llncs}
\usepackage[T1]{fontenc}
\usepackage{graphicx}
\usepackage{color}
\usepackage{makeidx}  % allows for index generation
\usepackage[colorlinks,linkcolor=blue, urlcolor=black]{hyperref}
\usepackage{url}
\usepackage{longtable}
\usepackage{multirow}
\usepackage{amssymb, amsmath, bm}
\usepackage{amsfonts, amssymb}
\usepackage{pifont}
\usepackage{mathtools}
\usepackage{mathrsfs}
\usepackage{booktabs}
\usepackage{cite}
\usepackage{dsfont}
\usepackage{makecell}
\usepackage[noend]{algpseudocode}
\usepackage{algorithmicx,algorithm}
\usepackage[misc]{ifsym} 
\usepackage{bbding}

\begin{document}

\title{Fourier Test-time Adaptation with Multi-level Consistency for Robust Classification}
\titlerunning{FTTA with Multi-level Consistency for Robust Classification}

\author{Yuhao Huang\inst{1,2,3}\thanks{Yuhao Huang and Xin Yang contribute equally to this work.}, Xin Yang\inst{1,2,3\star} \and Xiaoqiong Huang\inst{1,2,3} \and Xinrui Zhou\inst{1,2,3} \and Haozhe Chi\inst{4} \and Haoran Dou\inst{5} \and Xindi Hu\inst{6} \and Jian Wang\inst{7} \and Xuedong Deng\inst{8}\and Dong Ni\inst{1,2,3}\textsuperscript{(\Letter)}} 

% index{Huang, Yuhao}
% index{Yang, Xin}
% index{Huang, Xiaoqiong}
% index{Zhou, Xinrui}
% index{Chi, Haozhe}
% index{Dou, Haoran}
% index{Hu, Xindi}
% index{Wang, Jian}
% index{Deng, Xuedong}
% index{Ni, Dong}

\institute{
\textsuperscript{$1$}National-Regional Key Technology Engineering Laboratory for Medical Ultrasound, School of Biomedical Engineering, Health Science Center, Shenzhen University, China\\
\email{nidong@szu.edu.cn} \\
\textsuperscript{$2$}Medical Ultrasound Image Computing (MUSIC) Lab, Shenzhen University, China\\
\textsuperscript{$3$}Marshall Laboratory of Biomedical Engineering, Shenzhen University, China\\
\textsuperscript{$4$}ZJU-UIUC Institute, Zhejiang University, China\\
\textsuperscript{$5$}Centre for Computational Imaging and Simulation Technologies in Biomedicine (CISTIB), University of Leeds, UK\\ 
\textsuperscript{$6$}Shenzhen RayShape Medical Technology Co., Ltd, China\\
\textsuperscript{$7$}School of Biomedical Engineering and Informatics, Nanjing Medical University, China\\
\textsuperscript{$8$}The Affiliated Suzhou Hospital of Nanjing Medical University, China} 
% Center for Medical Ultrasound, 

\authorrunning{Huang et al.}
\maketitle              % typeset the header of the contribution

\begin{abstract}
Deep classifiers may encounter significant performance degradation when processing unseen testing data from varying centers, vendors, and protocols. Ensuring the robustness of deep models against these domain shifts is crucial for their widespread clinical application. In this study, we propose a novel approach called Fourier Test-time Adaptation (FTTA), which employs a dual-adaptation design to integrate input and model tuning, thereby jointly improving the model robustness. The main idea of FTTA is to build a reliable multi-level consistency measurement of paired inputs for achieving self-correction of prediction. Our contribution is two-fold. First, we encourage consistency in global features and local attention maps between the two transformed images of the same input. Here, the transformation refers to \textit{Fourier}-based input adaptation, which can transfer one unseen image into source style to reduce the domain gap. Furthermore, we leverage style-interpolated images to enhance the global and local features with learnable parameters, which can smooth the consistency measurement and accelerate convergence. Second, we introduce a regularization technique that utilizes style interpolation consistency in the frequency space to encourage self-consistency in the logit space of the model output. This regularization provides strong self-supervised signals for robustness enhancement. FTTA was extensively validated on three large classification datasets with different modalities and organs. Experimental results show that FTTA is general and outperforms other strong state-of-the-art methods. 


\end{abstract}

%======================= Intro =====================================
\section{Introduction}
\label{sec:intro}

\begin{figure*}[!h]
    \centering
    \includegraphics[width=0.9\linewidth]{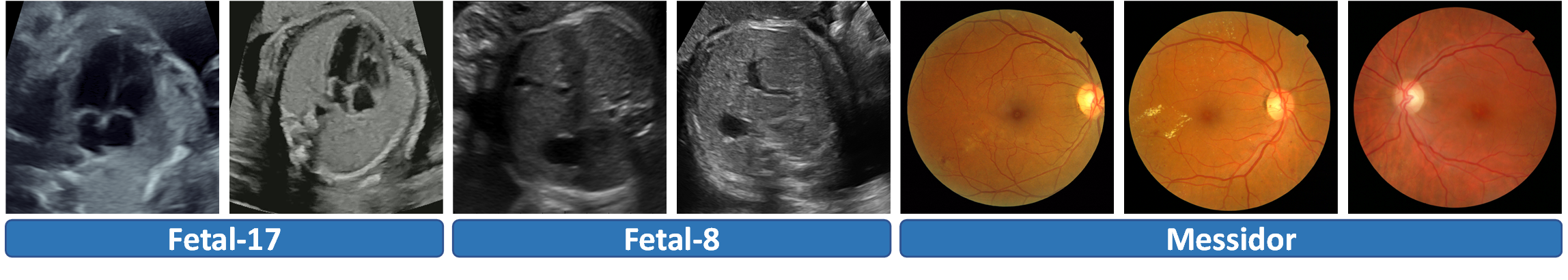}
    \caption{From left to right: 1) four-chamber views of heart from Vendor A\&B, 2) abdomen planes from Vendor C\&D, 3) fundus images with diabetic retinopathy of grade 3 from Center E-G. Appearance and distribution differences can be seen in each group.} 
    \label{fig:dataset}
\end{figure*}

Domain shift (see Fig.~\ref{fig:dataset}) may cause deep classifiers to struggle in making plausible predictions during testing~\cite{liu2021ttt++}.
This risk seriously limits the reliable deployment of these deep models in real-world scenarios, especially for clinical analysis.
Collecting data from the target domain to retrain from scratch or fine-tune the trained model is the potential solution to handle the domain shift risks.
However, obtaining adequate testing images with manual annotations is laborious and impracticable in clinical practice.
Thus, different solutions have been proposed to conquer the problem and improve the model robustness. \par

\textbf{Unsupervised Domain Adaptation (UDA)} refers to training the model with labeled source data and adapting it with target data without annotation~\cite{geng2011daml,tzeng2017adversarial,ren2018adversarial}.
Recently, \textit{Fourier} domain adaptation was proposed in~\cite{yang2020fda,zakazov2022feather}, with the core idea of achieving domain transfer by replacing the low-frequency spectrum of source data with that of the target one.
Although effective, they require obtaining sufficient target data in advance, which is challenging for clinical practice.

\textbf{Domain generalization (DG)} aims to generalize models to the unseen domain not presented during training. 
Adversarial learning-based DG is one of the most popular choices that require multi-domain information for learning domain-invariant representations~\cite{li2018domain, li2018deep}.
Recently, Liu et al.~\cite{liu2021feddg} proposed to construct a continuous frequency space to enhance the connection between different domains.
Atwany et al.~\cite{atwany2022drgen} imposed a regularization to reduce gradient variance from different domains for diabetic retinopathy classification. 
One drawback is that they require multiple types of source data for extracting rich features. 
Other alternatives proposed using only one source domain to perform DG~\cite{zhao2020maximum,fan2021adversarially}.
However, they still heavily rely on simulating new domains via various data augmentations, which can be challenging to control.

\textbf{Test-time Adaptation (TTA)} adapts the target data or pre-trained models during testing~\cite{huang2022online}. 
Test-time Training (TTT)~\cite{sun2020test} and TTT++~\cite{liu2021ttt++} proposed to minimize a self-supervised auxiliary loss. 
Wang et al.~\cite{wang2021tent} proposed the TENT framework that focused on minimizing the entropy of its predictions by modulating features via normalization statistics and transformation parameters estimation. 
Instead of batch input like the above-mentioned methods, Single Image TTA (SITA)~\cite{khurana2021sita} was proposed with the definition that having access to only one given test image once.
Recently, different mechanisms were developed to optimize the TTA including distribution calibration~\cite{ma2022test}, dynamic learning rate~\cite{yang2022dltta}, and normalizing flow~\cite{osowiechi2023tttflow}.
Most recently, Gao et al.~\cite{gao2023back} proposed projecting the test image back to the source via the source-trained diffusion models.
Although effective, these methods often suffer from the problems of unstable parameter estimation, inaccurate proxy tasks/pseudo labels, difficult training, etc. Thus, a simple yet flexible approach is highly desired to fully mine and combine information from test data for online adaptation.

In this study, we propose a novel framework called Fourier TTA (FTTA) to enhance the model robustness.
We believe that this is the first exploration of dual-adaptation design in TTA that jointly updates input and model for online refinement.
Here, one assumption is that a well-adapted model will get consistent outputs for different transformations of the same image.
Our contribution is two-fold.
First, we align the high-level features and attention regions of transformed paired images for complementary consistency at global and local dimensions.
We adopt the \textit{Fourier}-based input adaptation as the transformation strategy, which can reduce the distances between unseen testing images and the source domain, thus facilitating the model learning.
We further propose to smooth the hard consistency via the weighted integration of features, thus reducing the adaptation difficulties of the model.
Second, we employ self-consistency of frequency-based style interpolation to regularize the output logits. 
It can provide direct and effective hints to improve model robustness.
Validated on three classification datasets, we demonstrate that FTTA is general in improving classification robustness, and achieves state-of-the-art results compared to other strong TTA methods.

\begin{figure*}[!t]
    \centering
    \includegraphics[width=1.0\linewidth]{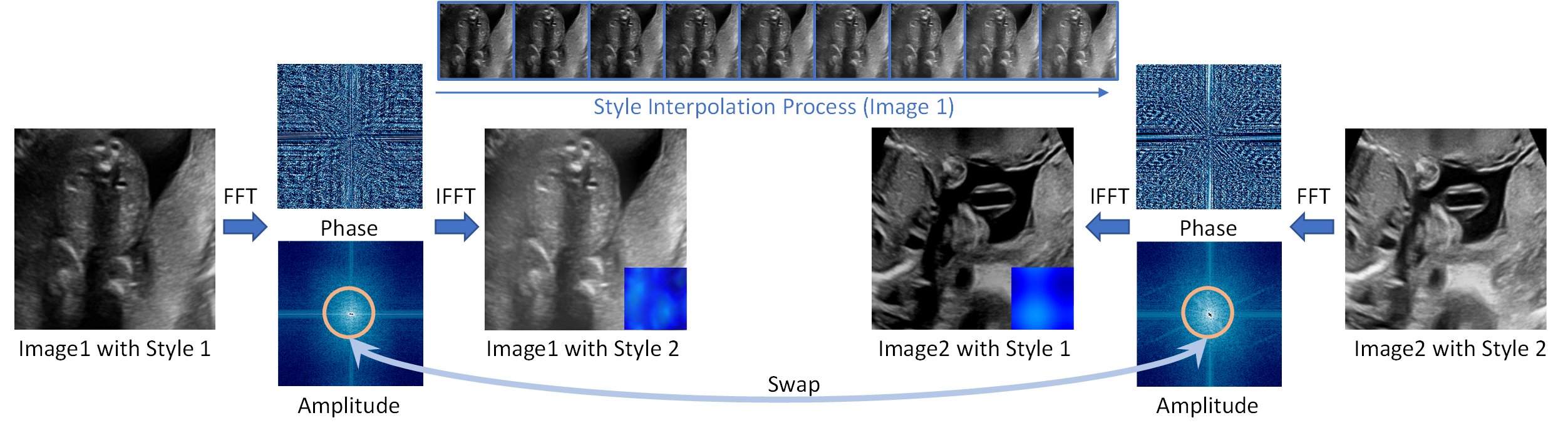}
    \caption{Illustration of the amplitude swapping between two images with different styles. Pseudo-color images shown in the right-down corner indicate the differences between images before and after amplitude swapping.} 
    \label{fig:fda}
\end{figure*}

%======================= Method ====================================
\section{Methodology}
\label{sec:method}
\begin{figure*}[!t]
    \centering
    \includegraphics[width=1.0\linewidth]{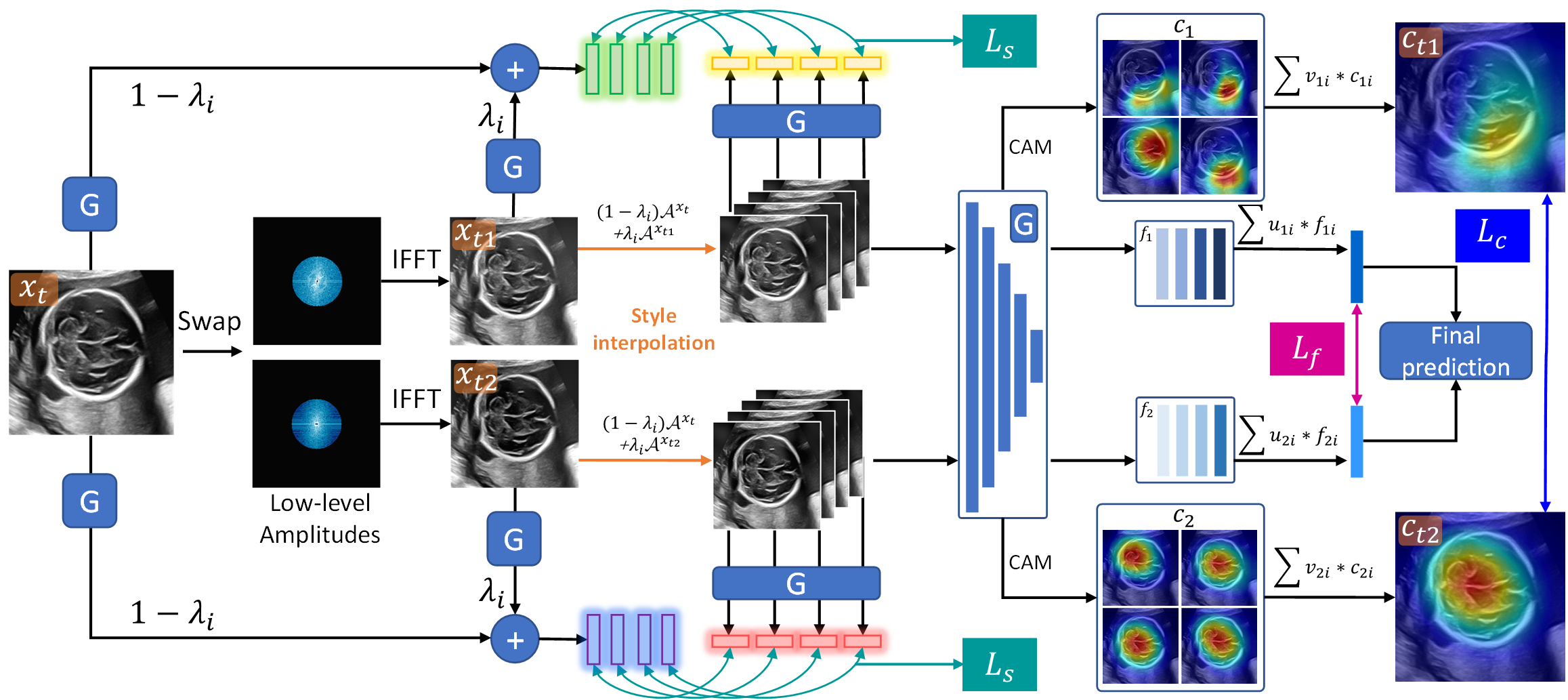}
    \caption{Pipeline of our proposed FTTA framework.} 
    \label{fig:TTA_testing}
\end{figure*}

Fig.~\ref{fig:TTA_testing} shows the pipeline of FTTA.
Given a trained classifier \textit{G}, FTTA first conducts $Fourier$-based input adaptation to transfer each unseen testing image $x_t$ into two source-like images ($x_{t1}$ and $x_{t2}$).
Then, using linear style interpolation, two groups of images will be obtained for subsequent smooth consistency measurement at global features ($L_{f}$) and local visual attention ($L_{c}$).
Furthermore, regularization in the logit space can be computed following the style interpolation consistency in the frequency space ($L_{s}$).
Finally, FTTA updates once based on the multi-consistency losses to output the final average prediction.

\textbf{Fourier-based Input Adaptation for Domain Transfer.}
Transferring unseen images to the known domain plays an important role in handling domain shift risks.
In this study, instead of learning on multiple domains, we only have access to one single domain of data during training.
Therefore, we need to utilize the limited information and find an effective way to realize the fast transfer from the unseen domain to the source domain.
Inspired by~\cite{yang2020fda,sharifzadeh2021ultrasound,zakazov2022feather}, we adopt the Fast Fourier Transform (FFT) based strategy to transfer the domain information and achieve input adaptation during testing.
Specifically, we transfer the domain information from one image to another by low-frequency amplitude ($\mathcal{A}$) swapping while keeping the phase components (see Fig.~\ref{fig:fda}).
This is because in \textit{Fourier} space, the low-frequency $\mathcal{A}$ encodes the style information, and semantic contents are preserved in $\mathcal{P}$~\cite{yang2020fda}.
Domain transfer via amplitude swapping between image $x_s$ to $x_{t}$ can be defined as:
\begin{equation}
\label{eq:A_obtain}
\mathcal{A}^{x_{t'}} = \left((1-\lambda) \mathcal{A}^{x_t}+\lambda \mathcal{A}^{x_{s}}\right) \circ \mathcal{M}+\mathcal{A}^{x_t} \circ (1-\mathcal{M}),
\end{equation}
where $\mathcal{M}$ is the circular low-pass filtering with radius $r$ to obtain the radial-symmetrical amplitude~\cite{zakazov2022feather}.
$\lambda$ aims to control the degree of style interpolation~\cite{liu2021feddg}, and it can make the transfer process continues (see Fig.~\ref{fig:fda}).
After inverse FFT~(IFFT, $\mathcal{F}^{-1}$), we can obtain an image $x_t'$ by $\mathcal{F}^{-1}(\mathcal{A}^{x_t'}, \mathcal{P}^{x_t})$.

Since one low-level amplitude represents one style, we have \textit{n} style choices. 
\textit{n} is the number of training data.
The chosen styles for input adaptation should be representative of the source domain while having significant differences from each other. 
Hence, we use the validation set to select the styles by first turning the whole validation data into the $n$ styles and calculating $n$ accuracy.
Then, styles for achieving $top$-$k$ performance are considered representative, and L2 distances between the $C_K^2$ pairs are computed to reflect the differences.

\textbf{Smooth Consistency for Global and Local Constraints.}
Building a reliable consistency measurement of paired inputs is the key to achieving TTA.
In this study, we propose global and local alignments to provide a comprehensive consistency signal for tuning the model toward robustness.
For \textit{global consistency}, we compare the similarity between high-level features of paired inputs.
These features encode rich semantic information and are therefore well-suited for assessing global consistency.
Specifically, we utilize hard and soft feature alignments via pixel-level L2 loss and distribution-level cosine similarity loss, to accurately compute the global feature loss $L_{f}$.
To ensure \textit{local consistency}, we compute the distances between the classification activation maps (CAMs) of the paired inputs.
It is because CAMs (e.g., Grad-CAM~\cite{selvaraju2017grad}) can reflect the local region the model focuses on when making predictions.
Forcing CAMs of paired inputs to be close can guide the model to optimize the attention maps and predict using the correct local region for refining the prediction and improving model robustness (see Fig.~\ref{fig:TTA_testing}, $c_{t1}$ is encouraged to be closer with $c_{t2}$ for local visual consistency).
Finally, the distances between two CAMs can be computed by the combination of L2 and JS-divergence losses.

Despite global and local consistency using single paired images can provide effective self-supervised signals for TTA in most cases, they may be difficult or even fail in aligning the features with a serious gap during testing.
This is because the representation ability of single-paired images is limited, and the hard consistency between them may cause learning and convergence difficulties.
For example, the left-upper CAMs of \textit{c}1 and \textit{c}2 in Fig.~\ref{fig:TTA_testing} are with no overlap. Measuring the local consistency between them is meaningless since JS divergence will always output a constant in that case. 
Thus, we first generate two groups of images, each with four samples, by style interpolation using different $\lambda$. Then, we fed them into the model for obtaining two groups of features.
Last, we propose learnable integration with parameters \textit{u} and \textit{v} to linearly integrate the global and local features.
This can enhance the feature representation ability, thus smoothing the consistency evaluation to accelerate the adaptation convergence.

\textbf{Style Consistency for Regularization on Logit Space.}
As described in the first half of Eq.~\ref{eq:A_obtain}, two low-level amplitudes (i.e., styles) can be linearly combined into a new one.
We propose to use this frequency-based style consistency to regularize the model outputs in logit space, which is defined as the layer before \textit{softmax}.
Thus, it is directly related to the model prediction.
A total of 8 logit pairs can be obtained (see Fig.~\ref{fig:TTA_testing}), and the loss can be defined as:
\begin{equation}
L_{s} = \bigg(\sum_{i = 1}^{2} \sum_{j = 1}^{4} ||(1-\lambda_{j})*y_{log}(x_{t})+\lambda_{j}*y_{log}(x_{ti})-y_{log}(x_{ij})||_{2}\bigg)/8,
\end{equation}
where $x_t$ and $x_{ti,i\in{1,2}}$ are the testing image and two transformed images after input adaptation.
$x_{ij}$ represents style-interpolated images controlled by $\lambda_{j}$. 
$y_{log}(\cdot)$ outputs the logits of the model.

%======================= Experimental Results ======================
\section{Experimental Results}
\label{sec:Experimental Results}
\begin{table}[!t]
  \centering
  \caption{Datasets split of each experimental group.}
    \setlength{\tabcolsep}{5mm}
    \begin{tabular}{ccccc}
    \toprule
          & Groups & Training  & Validation & Testing \\
    \midrule
    \multirow{2}[2]{*}{Fetal-17} & A2B   & 2622  & 1135  & 4970 \\
          & B2A   & 3472  & 1498  & 3757 \\
    \midrule
    \multirow{2}[2]{*}{Fetal-8} & C2D   & 3551  & 1529  & 5770 \\
          & D2C   & 4035  & 1735  & 5080 \\
    \midrule
    \multirow{3}[2]{*}{Messidor} & E2FG  & 279   & 121   & 800 \\
          & F2EG  & 278   & 122   & 800 \\
          & G2EF  & 279   & 121   & 800 \\
    \bottomrule
    \end{tabular}%
  \label{tab:dataset}%
\end{table}%

\textbf{Materials and Implementations.}
We validated the FTTA framework on three classification tasks, including one private dataset and two public datasets (see Fig.~\ref{fig:dataset}). 
Approved by the local IRB, the in-house \textit{Fetal-17} US dataset containing 8727 standard planes with gestational age (GA) ranging from $20$ to $24^{+6}$ weeks was collected.
It contains 17 categories of planes with different parts, including limbs (4), heart (4), brain (3), abdomen (3), face (2), and spine (1).
Four 10-year experienced sonographers annotated one classification tag for each image using the Pair annotation software package~\cite{liang2022sketch}.
\textit{Fetal-17} consists of two vendors (A\&B) and we conducted bidirectional experiments (A2B and B2A) for method evaluation.
The Maternal-fetal US dataset named \textit{Fetal-8} (GA: 18-40 weeks)~\cite{burgos2020evaluation}\footnote{\url{https://zenodo.org/record/3904280\#.YqIQvKhBy3A}} contains 8 types of anatomical planes including brain (3), abdomen (1), femur (1), thorax (1), maternal cervix (1), and others (1). 
Specifically, 10850 images from vendors ALOKA and Voluson (C\&D) were used for bidirectional validation (C2D and D2C).
Another public dataset is a fundus dataset named \textit{Messidor}, which contains 1200 images from 0-3 stage of diabetic retinopathy~\cite{decenciere2014feedback}\footnote{\url{https://www.adcis.net/en/third-party/messidor/}}.
It was collected from three ophthalmologic centers (E, F\&G) with each of them can treated as a source domain, allowing us to conduct three groups of experiments (E2FG, F2EG and G2EF).
Dataset split information is listed in Table~\ref{tab:dataset}.

We implemented FTTA in Pytorch, using an NVIDIA A40 GPU.
All images were resized to 256$\times$256, and normalized before input to the model.
For the fetal datasets, we used a 1-channel input, whereas, for the fundus dataset, 3-channel input was utilized.
During training, we augmented the data using common strategies including rotation, flipping and contrast transformation.
We selected ImageNet-pretrained \textit{ResNet-18}~\cite{he2016deep} as our classifier backbone and optimized it using the AdamW optimizer in 100 epochs.
For offline training, with batch size=196, the learning rate ($lr$) is initialized to 1e-3 and multiplied by 0.1 per 30 epochs. 
Cross-entropy loss is the basic loss for training.
We selected models with the best performance on validation sets to work with FTTA.
For online testing, we set the $lr$ equal to 5e-3, and $\lambda_{j,j=1,2,3,4}$ for style interpolation was set as 0.2, 0.4, 0.6, and 0.8, respectively.
We only updated the network parameters and learnable weights once based on the multi-level consistency losses function before obtaining the final predictions.

\begin{table}[!t]
  \centering
  \caption{Comparisons on Fetal-17 dataset. The best results are shown in bold.}
  \setlength{\tabcolsep}{1.7mm}
    \begin{tabular}{ccccccccc}
    \toprule
    \multirow{2}[2]{*}{Methods} & \multicolumn{4}{c}{Fetal-17: A2B} & \multicolumn{4}{c}{Fetal-17: B2A} \\
    \cmidrule(lr){2-5} \cmidrule(lr){6-9}      
    & Acc   & Pre   & Rec   & F1    & Acc   & Pre   & Rec   & F1 \\
    \midrule
    Upper-bound & 96.33  & 95.79  & 94.54  & 94.61  & 91.81  & 89.28  & 88.89  & 88.67  \\
    Baseline~\cite{he2016deep} & 61.25  & 64.57  & 59.46  & 57.83  & 63.51  & 60.60  & 61.60  & 57.50  \\
    \midrule
    TTT~\cite{sun2020test}   & 71.91  & 65.62  & 67.33  & 63.28  & 72.77  & 57.34  & 58.11  & 55.51  \\
    TTT++~\cite{liu2021ttt++} & 78.65  & 79.08  & 75.21  & 73.20  & 78.60  & 76.15  & 71.36  & 71.02  \\
    TENT~\cite{wang2021tent}  & 76.48  & 74.54  & 71.41  & 69.40  & 75.83  & 73.38  & 70.75  & 67.01  \\
    DLTTA~\cite{yang2022dltta} & 85.51  & 88.30  & 83.87  & 83.83  & 83.20  & 81.39  & 76.77  & 76.94  \\
    DTTA~\cite{gao2023back}  & 87.00  & 86.93  & 84.48  & 83.87  & 83.39  & 82.17  & 78.73  & 79.09  \\
    TTTFlow~\cite{osowiechi2023tttflow} & 86.66  & 85.98  & 85.38  & 84.96  & 84.08  & \textbf{85.41} & 76.99  & 75.70  \\
    \midrule
    FTTA-IA & 73.56  & 72.46  & 67.49  & 61.47  & 69.07  & 64.17  & 52.81  & 50.64  \\
    FTTA-C1 & 82.27  & 81.66  & 76.72  & 75.32  & 81.55  & 78.56  & 68.43  & 67.17  \\
    FTTA-C2 & 83.06  & 82.42  & 81.81  & 79.92  & 81.95  & 69.03  & 67.51  & 65.78  \\
    FTTA-C3 & 84.77  & 82.14  & 77.04  & 76.30  & 82.09  & 78.57  & 80.24  & 77.15  \\
    FTTA-C$^{*}$ & 80.70  & 82.52  & 77.24  & 78.18  & 79.24  & 80.64  & 74.42  & 74.00  \\
    FTTA-C & 88.93  & 89.43  & 82.40  & 82.96  & 84.91  & 80.91  & 75.64  & 76.52  \\
    Ours  & \textbf{91.02}  & \textbf{89.62}  & \textbf{89.74}  & \textbf{89.37}  & \textbf{87.41}  & 82.46  & \textbf{81.91}  & \textbf{81.15}  \\
    \bottomrule
    \end{tabular}%
  \label{tab:Fetal-17}%
\end{table}%

\begin{table}[!t]
  \centering
  \caption{Comparisons on Fetal-8 and MESSDIOR datasets.}
  \setlength{\tabcolsep}{1.2mm}
    \begin{tabular}{ccccccccccc}
    \toprule
    \multirow{2}[4]{*}{Methods} & \multicolumn{2}{c}{C2D} & \multicolumn{2}{c}{D2C} & \multicolumn{2}{c}{E2FG} & \multicolumn{2}{c}{F2EG} & \multicolumn{2}{c}{G2EF} \\
    \cmidrule(lr){2-3} \cmidrule(lr){4-5} \cmidrule(lr){6-7} \cmidrule(lr){8-9} \cmidrule(lr){10-11}
    & Acc   & F1    & Acc   & F1    & Acc   & F1    & Acc   & F1    & Acc   & F1 \\
    \midrule
    Upper-bound & 87.49  & 85.14  & 94.05  & 85.24  & 60.91  & 53.46  & 66.26  & 57.63  & 62.14  & 53.34  \\
    Baseline & 67.68  & 65.93  & 79.92  & 72.09  & 47.62  & 37.09  & 31.13  & 31.30  & 41.25  & 39.41  \\
    Ours  & 82.13  & 76.89  & 91.87  & 73.19  & 59.26  & 43.98  & 57.43  & 37.26  & 58.02  & 45.46  \\
    \bottomrule
    \end{tabular}%
  \label{tab:Fetal8-FUNDUS}%
\end{table}%

\textbf{Quantitative and Qualitative Analysis.}
We evaluated the classification performance using four metrics including Accuracy (\textit{Acc},~\%), Precision (\textit{Pre},~\%), Recall (\textit{Rec},~\%), and F1-score (\textit{F1},~\%).
Table~\ref{tab:Fetal-17} compares the FTTA (\textit{Ours}) with seven competitors including the \textit{Baseline} without any adaptation and six state-of-the-art TTA methods.
\textit{Upper-bound} represents the performance when training and testing on the target domain.
It can be seen from \textit{Upper-bound} and \textit{Baseline} that all the metrics have serious drops due to the domain shift.
\textit{Ours} achieves significant improvements on \textit{Baseline}, and outperforms all the strong competitors in terms of all the evaluation metrics, except for the \textit{Pre} in Group B2A.
It is also noted that the results of \textit{Ours} are approaching the \textit{Upper-bound}, with only 5.31\% and 4.40\% gaps in \textit{Acc}.

We also perform ablation studies on the \textit{Fetal-17} dataset in the last 7 rows of Table~\ref{tab:Fetal-17}.
\textit{FTTA-IA} denotes that without model updating, only input adaptation is conducted.
Four experiments are performed to analyze the contribution of three consistency measurements (\textit{-C1}, \textit{-C2}, and \textit{-C3} for global features, local CAM, and style regularization, respectively), and also the combination of them (\textit{-C}).
They are all equipped with the input adaptation for fair comparisons.
\textit{FTTA-C$^{*}$} indicates replacing the Fourier-based input adaptation with 90$^{\circ}$ rotation to augment the test image for consistency evaluation.
Different from \textit{FTTA-C}, \textit{Ours} integrates learnable weight groups to smooth consistency measurement.
Experiences show that the naive Fourier input adaptation in \textit{FTTA-IA} can boost the performance of \textit{Baseline}.
The three consistency variants improve the classification performance respectively, and combining them together can further enhance the model robustness.
Then, the comparison between \textit{FTTA-C} and \textit{Ours} validates the eﬀectiveness of the consistency smooth strategy.

\begin{figure*}[!t]
    \centering
    \includegraphics[width=0.9\linewidth]{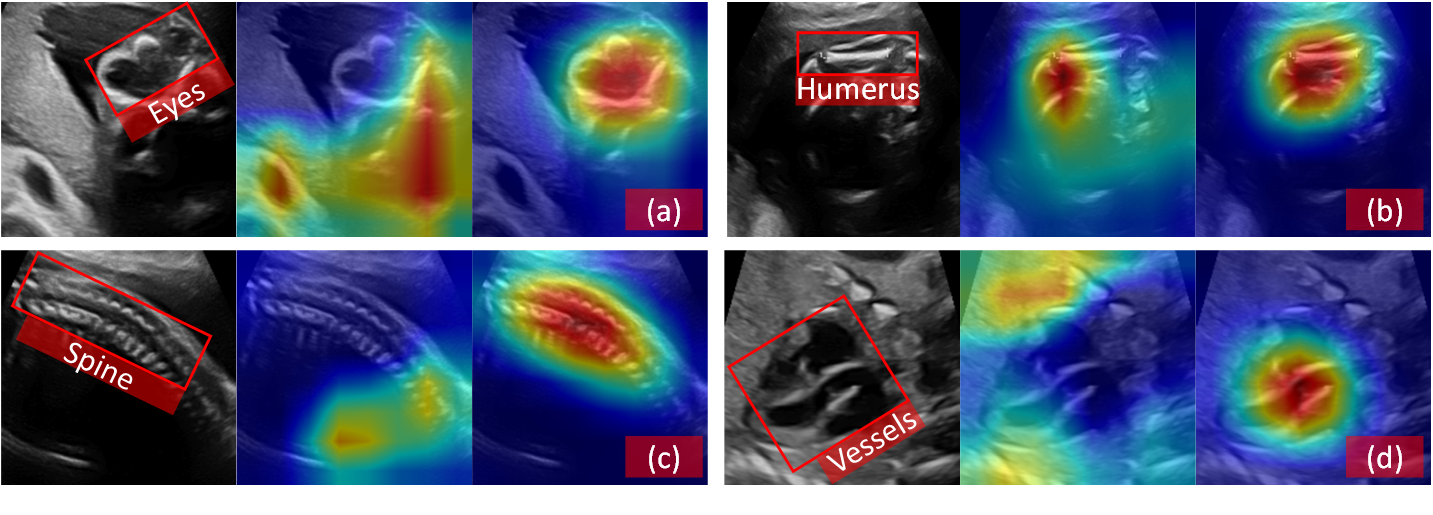}
    \caption{Four typical cases in Fetal-17 dataset: (a) Axial orbit and lenses, (b) Humerus plane, (c) Sagittal plane of the spine, and (d) Left ventricular outflow tract view.}
    \label{fig:CAM_result}
\end{figure*}

Table~\ref{tab:Fetal8-FUNDUS} reports the results of FTTA on two public datasets.
We only perform methods including \textit{Upper-bound}, \textit{Baseline}, and \textit{Ours} with evaluation metrics \textit{Acc} and \textit{F1}.
Huge domain gaps can be observed by comparing \textit{Upper-bound} and \textit{Baseline}.
All five experimental groups prove that our proposed FTTA can boost the classification performance over \textit{baseline}, and significantly narrow the gaps between \textit{upper-bound}.
Note that \textit{MESSDIOR} is a challenging dataset, with all the groups having low Upper-bounds.
Even for the multi-source DG method, \textit{Messidor} only achieves 66.70$\%$ accuracy~\cite{atwany2022drgen}.
For the worst group (F2EG), \textit{Acc} drops 35.13$\%$ in the testing sets.
However, the proposed FTTA can perform a good adaptation and improve 26.30$\%$ and 5.96$\%$ in \textit{Acc} and \textit{F1}. 

Fig.~\ref{fig:CAM_result} shows the CAM results obtained by \textit{Ours}.
The red boxes denote the key regions, like the \textit{eyes} in (a), which were annotated by sonographers and indicate the region-of-interest (ROI) with discriminant information.
We consider that if one model can focus on the region having a high overlap with the ROI box, it has a high possibility to be predicted correctly.
The second columns visualize the misclassified results before adaptation.
It can be observed via the CAMs that the focus of the model is inaccurate.
Specifically, they spread dispersed on the whole image, overlap little with the ROI, or with    low prediction confidence.
After TTA, the CAMs can be refined and close to the ROI, with prediction corrected.

%======================= Conclusion ================================
\section{Conclusion}
\label{sec:conclusion}
In this study, we proposed a novel and general FTTA framework to improve classification robustness.
Based on Fourier-based input adaptation, FTTA is driven by the proposed multi-level consistency, including smooth global and local constraints, and also the self-consistency on logit space.
Extensive experiments on three large datasets validate that FTTA is effective and efficient, achieving state-of-the-art results over strong TTA competitors.
In the future, we will extend the FTTA to segmentation or object detection tasks.

%======================= Acknowledgement ===========================
\subsubsection{Acknowledgement.} 
This work was supported by the grant from National Natural Science Foundation of China (Nos. 62171290, 62101343), Shenzhen-Hong Kong Joint Research Program (No. SGDX20201103095613036), and Shenzhen Science and Technology Innovations Committee (No. 20200812143441001).

\bibliographystyle{splncs04}
\bibliography{reference}

\end{document}

% \title{Supplementary Material\\ \normalsize{Paper ID\# 1509}}
\title{Supplementary Material}
\author{ }

\institute{}

\maketitle

\begin{table}[htbp]
  \centering
  \caption{Details of Fetal-17 dataset.}
  \setlength{\tabcolsep}{6.9mm}
    \begin{tabular}{cccc}
    \toprule
    \multirow{2}[4]{*}{ID} & \multirow{2}[4]{*}{Class Name} & \multicolumn{2}{c}{Numbers} \\
\cmidrule{3-4}          &       & A     & B \\
    \midrule
    0     & Coronal lips and nose & 127   & 283 \\
    1     & Trans-ventricular plane & 214   & 286 \\
    2     & Forearm & 91    & 440 \\
    3     & Humerus plane & 245   & 325 \\
    4     & Femur plane & 408   & 426 \\
    5     & Sagittal plane of the spine & 465   & 530 \\
    6     & Leg bones & 49    & 148 \\
    7     & Umbilical cord insertion & 116   & 444 \\
    8     & Trans-thalamic plane & 193   & 288 \\
    9     & Three vessels and trachea view & 270   & 308 \\
    10    & The axial plane of the abdomen & 222   & 248 \\
    11    & Axial kidneys & 160   & 203 \\
    12    & Axial orbit and lenses & 232   & 263 \\
    13    & Four chamber view of heart & 414   & 353 \\
    14    & Trans-cerebellar plane & 224   & 151 \\
    15    & Right ventricular outflow tract view & 243   & 57 \\
    16    & Left ventricular outflow tract view  & 84    & 217 \\
    \midrule
    \multicolumn{2}{c}{Total} & 3757  & 4970 \\
    \bottomrule
    \end{tabular}%
  \label{tab:Fetal-17}%
\end{table}%

\begin{table}[htbp]
  \centering
  \caption{Details of Fetal-8 dataset.}
  \setlength{\tabcolsep}{10mm}
    \begin{tabular}{cccc}
    \toprule
    \multirow{2}[4]{*}{ID} & \multirow{2}[4]{*}{Class Name} & \multicolumn{2}{c}{Numbers} \\
\cmidrule{3-4}          &       & C     & D \\
    \midrule
    0     & Fetal-abdomen & 212   & 366 \\
    1     & Fetal-femur & 294   & 623 \\
    2     & Fetal-thorax & 291   & 1324 \\
    3     & Maternal-cervix & 610   & 897 \\
    4     & Trans-cerebellum & 134   & 492 \\
    5     & Trans-thalamic & 360   & 1072 \\
    6     & Trans-ventricular & 112   & 408 \\
    7     & Other & 3067  & 588 \\
    \midrule
    \multicolumn{2}{c}{Total} & 5080  & 5770 \\
    \bottomrule
    \end{tabular}%
  \label{tab:Fetal-8}%
\end{table}%

\begin{table}[htbp]
  \centering
  \caption{Details of MESSDIOR dataset.}
  \setlength{\tabcolsep}{10mm}
    \begin{tabular}{ccccc}
    \toprule
    \multirow{2}[4]{*}{ID} & \multirow{2}[4]{*}{Class Name} & \multicolumn{3}{c}{Numbers} \\
\cmidrule{3-5}          &       & E     & F     & G \\
    \midrule
    0     & Normal & 151   & 186   & 209 \\
    1     & DR-Grade1 & 30    & 71    & 52 \\
    2     & DR-Grade2 & 70    & 91    & 86 \\
    3     & DR-Grade3 & 149   & 52    & 53 \\
    \midrule
    \multicolumn{2}{c}{Total} & 400   & 400   & 400 \\
    \bottomrule
    \end{tabular}%
  \label{tab:MESSDIOR}%
\end{table}%